\documentclass{ieeeaccess}
\usepackage{cite}
\usepackage{amsmath,amssymb,amsfonts}
\usepackage{algorithmic}
\usepackage{graphicx}
\usepackage{textcomp}
\usepackage{multirow, multicol, booktabs, makecell}
\usepackage{float}
\usepackage{amssymb}
\renewcommand{\thetable}{\Roman{table}}
\usepackage{bm}
\usepackage{diagbox}
\usepackage[hidelinks]{hyperref}

\makeatletter
\AtBeginDocument{\DeclareMathVersion{bold}
\SetSymbolFont{operators}{bold}{T1}{times}{b}{n}
\SetSymbolFont{NewLetters}{bold}{T1}{times}{b}{it}
\SetMathAlphabet{\mathrm}{bold}{T1}{times}{b}{n}
\SetMathAlphabet{\mathit}{bold}{T1}{times}{b}{it}
\SetMathAlphabet{\mathbf}{bold}{T1}{times}{b}{n}
\SetMathAlphabet{\mathtt}{bold}{OT1}{pcr}{b}{n}
\SetSymbolFont{symbols}{bold}{OMS}{cmsy}{b}{n}
\renewcommand\boldmath{\@nomath\boldmath\mathversion{bold}}}
\makeatother

\def\BibTeX{{\rm B\kern-.05em{\sc i\kern-.025em b}\kern-.08em
    T\kern-.1667em\lower.7ex\hbox{E}\kern-.125emX}}

\begin{document}

\title{Deep Learning-Based Noninvasive Screening of Type 2 Diabetes with Chest X-ray Images and Electronic Health Records}
\author{\uppercase{Sanjana Gundapaneni},
\uppercase{Zhuo Zhi}, and
\uppercase{Miguel Rodrigues}}
\address[]{Department of Electronic and Electrical Engineering, University College London, WC1E 6BT London, U.K.}

\tfootnote{\color{red}This work has been submitted to the IEEE for possible publication. Copyright may be transferred without notice, after which this version may no longer be accessible.}

\markboth
{S. Gundapaneni \headeretal: Deep Learning-Based Noninvasive Screening of T2DM with CXR images and EHRs}
{S. Gundapaneni \headeretal: Deep Learning-Based Noninvasive Screening of T2DM with CXR images and EHRs}

\corresp{Corresponding author: Sanjana Gundapaneni (\href{mailto:sanjana.gundapaneni.21@ucl.ac.uk}{sanjana.gundapaneni.21@ucl.ac.uk})}

\begin{abstract}
The imperative for early detection of type 2 diabetes mellitus (T2DM) is challenged by its asymptomatic onset and dependence on suboptimal clinical diagnostic tests, contributing to its widespread global prevalence. While research into noninvasive T2DM screening tools has advanced, conventional machine learning approaches remain limited to unimodal inputs due to extensive feature engineering requirements. In contrast, deep learning models can leverage multimodal data for a more holistic understanding of patients’ health conditions. However, the potential of chest X-ray (CXR) imaging, one of the most commonly performed medical procedures, remains underexplored. This study evaluates the integration of CXR images with other noninvasive data sources, including electronic health records (EHRs) and electrocardiography signals, for T2DM detection. Utilising datasets meticulously compiled from the MIMIC-IV databases, we investigated two deep fusion paradigms: an early fusion-based multimodal transformer and a modular joint fusion ResNet-LSTM architecture. The end-to-end trained ResNet-LSTM model achieved an AUROC of 0.86, surpassing the CXR-only baseline by 2.3\% with just 9863 training samples. These findings demonstrate the diagnostic value of CXRs within multimodal frameworks for identifying at-risk individuals early. Additionally, the dataset preprocessing pipeline has also been released to support further research in this domain (available at \href{https://github.com/san-635/t2dm-cxr-ehr}{\underline{https://github.com/san-635/t2dm-cxr-ehr}}).
\end{abstract}

\titlepgskip=-21pt

\maketitle

\section{Introduction}
\PARstart{D}{iabetes} mellitus has long been described as a `silent killer' by the medical community \cite{I2001killer} due to its insidious impact on both microvascular and macrovascular systems, which raises the risk of hypertension, cardiovascular disease, chronic kidney disease, and other life-threatening comorbidities \cite{I2019cohort}. However, more concerningly, it has also emerged as an accelerating epidemic over the past few decades. In 2007, the International Diabetes Federation estimated that 7.1\% of the global population, aged between 20 and 79, would suffer from the chronic disease by 2025 \cite{I2007countries}. Yet, this percentage had already escalated to 10.5\% by 2021 \cite{I2022for}, despite preventive measures and heightened global awareness. Consequently, the prominence of Type 2 diabetes mellitus (T2DM), the most common variant, continues to increase. T2DM inhibits the body's ability to use insulin effectively, thus causing insulin resistance and high blood glucose levels.

The current clinical diagnosis of T2DM primarily relies on blood tests \cite{I2023diabetes}, where classification is based on fasting blood glucose or HbA1c thresholds. Although these tests can be conclusive for symptomatic patients, they present several limitations. Firstly, the procedures are invasive and some require fasting prior to administration. Additionally, the likelihood of false positives is substantial because HbA1c concentrations can be affected by factors such as ethnic backgrounds and other medical conditions, like sickle cell anaemia. More importantly, reliance on such thresholds can result in delayed detection, as the onset of T2DM is often asymptomatic \cite{I2004Diabetes}, and disease characteristics may only be identified once they have already progressed to a more severe stage \cite{I2023late}.

Furthermore, in developing countries, frequent testing is hindered by inadequate resources and a shortage of trained staff. As a result, efforts are often confined to high-risk patients identified by a narrow set of biomarkers, such as high Body Mass Index (BMI) or waist circumference. This approach can lead to certain demographics being overlooked \cite{I2024to}, particularly among younger populations, women, and those who appear mostly healthy. Consequently, a significant number of cases go undetected \cite{I2021edition}, preventing early care and intervention efforts, such as remission programmes. These gaps in the current diagnostic framework necessitate accurate and noninvasive T2DM screening tools.

With the advent of artificial intelligence (AI) in healthcare, significant research has focused on developing predictive models for T2DM and associated conditions \cite{R2021review}\cite{R2022review}. These models predominantly employ machine learning (ML) algorithms, including logistic regression (LR), support vector machines (SVMs), random forests (RFs), and ensemble models. Their performance has been promising, owing to the models' ability to learn from large amounts of patient information and produce insightful outputs. However, traditional ML models face a major limitation: they require manual feature extraction from high-dimensional data \cite{I2022Directions}. This restricts their application to unimodal inputs—such as electronic health records (EHRs) or noninvasive measurements like photoplethysmography (PPG) \cite{I2023Features} or electrocardiography (ECG) signals \cite{I2023electrocardiogram}. Despite the wealth of information available in these data sources, the unimodal approach fails to consider the patients' comprehensive health history, which is essential given the multifactorial nature of T2DM.

In response to this impediment, several deep learning (DL)-based screening approaches utilising multimodal data have been explored in recent years. These typically integrate single-point measurements of common risk factors with either omics data or biomedical imaging modalities \cite{I2023prediction}, with the latter predominantly relying on convolutional encoders. However, the range of image data incorporated in these studies has been comparatively limited. While abdominal computed tomography (CT) scans \cite{I2020Records} and retinal images \cite{I2022photographs}\cite{I2021images} have been more widely studied, chest X-rays (CXRs) remain underused. This is notable, as CXRs are the most commonly requested and performed medical imaging procedure \cite{I2023Release}, and recent DL research \cite{I2023radiographs} has demonstrated their prospective use in T2DM detection. Particularly, the utility of CXRs is linked to the identification of mediastinal fat deposits—accumulations of body fat within the central thoracic cavity (mediastinum) \cite{I2004Hypertension}—the size of which has been correlated to insulin resistance and T2DM in several medical case reports \cite{I2008report}\cite{I2012patient}.

Addressing the need for population-level, multimodal, noninvasive screening tools and evaluating the potential of CXRs in this context, this paper employs publicly available time series EHR data and CXR images for T2DM prediction. To the best of our knowledge, this is the first work to explore this particular combination. Furthermore, the inclusion of 12-lead ECG signals is examined to leverage the established link between T2DM and cardiovascular complications. Specifically, early signs of these complications often manifest as subtle subclinical abnormalities in ECGs \cite{I2024diabetes}, which are not discernible in EHR data, making them a valuable modality for enhancing the diagnostic capabilities of this study.

Two state-of-the-art DL classification models, following distinct multimodal fusion paradigms, are investigated for this task. The first approach, a joint fusion architecture, is represented by a hybrid Residual Network–Long Short-Term Memory (ResNet-LSTM) model with a modular encoder structure. The second approach involves a multimodal transformer that adopts an early fusion strategy. While the ResNet-LSTM model extends existing multimodal prediction frameworks by accommodating time series EHR data, transformer architectures remain relatively unexplored for T2DM diagnosis, despite being widely used in other medical downstream tasks, such as radiology report summarisation \cite{I2024Tasks}\cite{I2024Review}. By leveraging a self-attention mechanism, multimodal transformers can encapsulate underlying interactions between modalities \cite{I2023Survey}, enabling diagnostic procedures analogous to those followed by clinical professionals. The two key contributions of this paper are summarised below:
\begin{enumerate}
    \item We enhanced and extended the preprocessing pipeline proposed in \cite{I2019data} and constructed datasets comprising two diagnostically relevant modality combinations—(1) EHR and CXR, and (2) EHR, CXR, and ECG—using data from the Medical Information Mart for Intensive Care (MIMIC)-IV databases. The relevant code has been made publicly available, encouraging and facilitating future multimodal T2DM research.
    \item We demonstrated the effectiveness of combining time series EHR data and CXR images for predicting T2DM using different multimodal fusion paradigms. Our best-performing ResNet-LSTM model achieved a 2.3\% improvement in the Area Under the Receiver Operating Characteristic (AUROC) metric compared to the baseline CXR-only DL model \cite{I2023radiographs}, with significantly fewer training samples. This improvement increased to 2.6\% with the addition of ECG data.
\end{enumerate}

The rest of the paper is organised as follows: Section \ref{sec: related} provides an overview of existing studies at the intersection of T2DM screening and machine/deep learning. Section \ref{sec: methods} details the dataset preparation process and model architectures. Section \ref{sec: exp+results} presents the experimental setup, along with the results and discussion of the conducted experiments, including ablation studies. Finally, Section \ref{sec: conclusion} summarises the findings of this paper and discusses its limitations.

\section{Related Work}
\label{sec: related}
\subsection{Risk-scoring systems for T2DM}
Deterministic risk-scoring systems have been widely used to identify high-risk individuals \cite{R2011review}\cite{R2022review}. These methods often rely on biomarkers such as sociodemographic factors, waist circumference, and lifestyle indicators like medication and dietary habits to produce a quantitative risk level based on predefined thresholds. However, most such data-driven scales are tailored to specific ethnic or geographic populations, as their features and thresholds are customised for these groups, thereby limiting their applicability in broader settings.

\subsection{Unimodal conventional ML for T2DM prediction}
A significant body of research on T2DM detection focuses on applying conventional supervised ML algorithms, with studies often comparing multiple models. For instance, \cite{R2017records} evaluates LR, k-nearest neighbours (k-NN), Naïve Bayes (NB), SVM, J48 decision tree (DT), and RF classifiers using a regionally centralised EHR repository from Shanghai, China. Through several levels of abstraction, EHR feature sets with 107, 33, and 5 features are curated. The study finds that extensive feature engineering consistently yields superior performance across all classifiers, highlighting its critical role in traditional ML frameworks. Another study \cite{R2017paradigm} trains NB, Linear Discriminant Analysis (LDA), Quadratic Discriminant Analysis (QDA), and Gaussian Process Classifier (GPR) models on the Pima Indian Diabetic Database (PIDD), which contains eight EHR features for 768 adult females, including oral glucose tolerance test (OGTT) results. Although the kernel-based GPR achieves the highest accuracy of 0.820, PIDD's gender-specific composition limits its generalisability. Similar prediction limitations are observed in \cite{R2017study}, \cite{R2019Risks}, and \cite{R2019Techniques}, where training data is sourced from specific age groups, further emphasising the need for diverse and well-balanced datasets. Beyond EHRs, PPG and ECG waveforms have been explored using models such as extreme gradient boosting (XGBoost) \cite{I2023Features}\cite{I2023electrocardiogram}.

\subsection{Unimodal DL for T2DM prediction}
Given the ability of neural networks to process large datasets without substantial manual feature extraction, their application to EHR data has gained popularity \cite{R2021ReviewIEEE}. Studies such as \cite{R2020dataset} demonstrate the effectiveness of simpler algorithms like multilayer perceptrons (MLP), with a two-hidden-layer MLP achieving an accuracy of 0.981 on the PIDD, outperforming traditional models like DT and NB. More advanced DL architectures have also been studied, as done by the researchers of \cite{R2020diabetes}, where convolutional neural network (CNN), LSTM, joint CNN-LSTM, and hybrid convolutional-LSTM (Conv-LSTM) frameworks achieve accuracies of 0.887, 0.909, 0.921, and 0.973, respectively. Notably, they apply the RF-based Boruta algorithm for feature selection, revealing that the Conv-LSTM model achieves comparable performance using only the five most impactful features, owing to the LSTM network's ability to capture long-range relationships.

While accuracy is a common evaluation metric, it is often unreliable for imbalanced datasets like the PIDD, which includes only 31.5\% diabetic samples. To address this problem and leverage the enhanced learning capacity of DL models trained on larger datasets, authors of \cite{R2021augmentation} employ variational autoencoder (VAE)-based oversampling and a sparse autoencoder (SAE)-based feature augmentation, reporting an accuracy of 0.923 with a CNN and 0.857 with a single-hidden-layer MLP. Similarly, \cite{R2019records} explores the Synthetic Minority Oversampling Technique (SMOTE) on the highly imbalanced Practice Fusion dataset, obtaining improved sensitivity with a wide-and-deep architecture.

Furthermore, other notable methodologies include time series modelling \cite{R2018Models} and the adoption of multi-headed self-attention (MHSA) for analysing patients' past diagnoses \cite{R2023Records}, illustrating the breadth of DL techniques applied to T2DM prediction.

\subsection{Multimodal DL for T2DM prediction}
However, the far-reaching impacts of T2DM and its broad array of risk indicators necessitate an analysis that extends beyond EHR data alone. Motivated by this, many studies now incorporate additional clinical modalities, ranging from omics to biomedical imaging data. These inputs are typically combined with static EHR features, i.e., single-point measurements of risk factors such as age, biological sex, and BMI, through various fusion strategies.

Metabolomics, primarily explored within early fusion frameworks, is often integrated into conventional ML models due to its compatibility with EHR data structures \cite{R2017diabetes}\cite{R2024Chinese}\cite{R2022Chinese}. Consequently, feature selection methods, such as GreedyRLS, Boruta, and other RF-based algorithms, are required to handle the high dimensionality of metabolite profiles. For instance, \cite{R2017diabetes} implements multivariate LR with regularised least squares (RLS) to classify T2DM risk, using GreedyRLS to identify the optimal metabolites. The study obtains AUROCs of 0.680 with only EHR biomarkers, 0.770 with all 568 metabolites, and 0.780 with optimal multimodal inputs, highlighting the predictive advantage of combining modalities. Another study \cite{R2022Chinese} integrates genetic risk scores (GRS), clinical risk factors, and the five most impactful RF-derived metabolites, yielding a remarkable AUROC of 0.960, compared to 0.798 with clinical features alone and 0.923 using the five metabolites. Despite their demonstrated potential for T2DM diagnosis, the extraction of metabolomic biomarkers requires blood serum or plasma samples, rendering these approaches invasive.

Alternatively, retinal imaging \cite{I2022photographs}\cite{I2021images}\cite{R2022Network} and CT scans \cite{I2020Records}\cite{R2022Learning} have been investigated to develop noninvasive multimodal screening tools. Retinal images, in particular, are increasingly explored for their ability to detect signs of diabetic retinopathy and predict cardiovascular biomarkers, such as HbA1c and blood pressure \cite{R2018learning}. These studies predominantly utilise convolutional image encoders, such as ResNet variants and U-Net, to harness the spatial information captured by CNNs. Authors in \cite{I2021images} employ a pre-trained ResNet50 as a retinal image encoder, concatenating the extracted feature representations with clinical risk factors before feeding them to a three-hidden-layer MLP for joint learning. This architecture significantly improves T2DM detection, achieving an AUROC of 0.845, surpassing a baseline RF model with EHR data alone (AUROC = 0.762). Similarly, \cite{I2022photographs} uses ResNet18 to extract features from retinal images sourced from the UK Biobank database, and combines these with seven static risk-factor features to develop logistic models. The study reports AUROCs of 0.844 with multimodal data, 0.810 using only risk factors, and 0.731 with image-only models. These findings establish the viability of joint fusion designs involving modality-specific encoders as a prospective approach for multimodal T2DM research.

Similar architectures are explored in \cite{I2020Records}, which studies the fusion of abdominal CT scans and traditional biomarkers. Specifically, models are trained using data collected at least one year before clinical diagnosis, enabling the prediction of T2DM onset a year in advance. The EHR features include demographic data, pancreas volume extracted via a 3D U-Net abdominal segmentation model, visceral and subcutaneous fat volumes and distributions derived through fuzzy C-means clustering, and blood glucose test results, all projected linearly. CT scan images are processed using an unsupervised body-part regression algorithm to extract pancreas slices, which are then input to a simple CNN. Features from both modalities are fused before a binary classifier, forming a modular end-to-end framework. The multimodal model produces AUROC improvements of 8.43\% and 5.01\% over EHR-only and CT-only models, respectively. Contrarily, \cite{R2022Learning} implements a multivariate LR model that analyses five clinical risk factors alongside 23 CT-derived features extracted via a 3D U-Net. However, despite the computational expense of DL-based feature extraction, this hybrid approach yields only a modest AUROC of 0.680, thereby demonstrating the superiority of end-to-end DL methods for multimodal integration.

\subsection{CXR-based T2DM prediction}
The use of abdominal CT scans in this research area is relatively recent. Likewise, while retinal imaging has long been studied for detecting diabetic retinopathy \cite{R2020review}\cite{R2017DiabetesJAMA}, a visual impairment condition that develops years after disease onset \cite{R2009analysis}, its direct application in T2DM diagnosis is a newer focus. However, early detection of the disease could mitigate progression to irreversible retinopathy \cite{R2018Diabetes}, reinforcing the need for such T2DM-focused studies. Building on these exploratory directions in new biomedical imaging, \cite{I2023radiographs} investigates the potential of CXR images for T2DM screening using a ResNet34-based classifier. The study further employs occlusion-based explainable AI methods to produce visualisations, illustrating the relevance of CXRs for this task. Additionally, a hybrid approach integrates the CNN's predictions with an LR model alongside age, biological sex, BMI, ethnicity, language preference, and social deprivation index. While the CXR-only model achieves an AUROC of 0.840, the hybrid CNN-LR model attains 0.850.

This limited improvement can be attributed to the ineffective incorporation of CXR data. In particular, DL techniques are confined to feature extraction, while the LR model performs modality integration, potentially failing to capture interactions between the two modalities. Moreover, relying solely on prediction scores from ResNet34 disregards critical information within the CXR images. To address these shortcomings, this paper explores end-to-end DL models for more effective integration of CXR data in multimodal settings.

\section{Methods}
\label{sec: methods}
A labelled retrospective dataset, \textit{$\mathbf{D}_{\boldsymbol{E+C+G}}$}, is constructed for this study by integrating data from three clinical modalities: EHR (\(E\)), CXR (\(C\)), and ECG (\(G\)). The EHR-CXR-only dataset for this study, \textit{$\mathbf{D}_{\boldsymbol{E+C}}$}, is then derived by excluding the ECG modality to facilitate later comparability. Additionally, a pipeline for compiling a separate \textit{$\mathbf{D}_{\boldsymbol{E+C}}$} dataset, completely independent of ECG data, is provided. Data for these modalities are sourced from the MIMIC-IV version 2.2 \cite{MIMIC-IV}\cite{MIMIC-IV-orig}, MIMIC-CXR-JPG version 2.0.0 \cite{MIMIC-CXR-JPG}\cite{MIMIC-CXR-JPG-orig}, and MIMIC-IV-ECG version 1.0 \cite{MIMIC-IV-ECG} databases, respectively. Access to these datasets was granted through the PhysioNet platform \cite{PhysioNet} after completing the necessary training and signing the Data Use Agreements.

The preprocessing pipelines adopted to convert the raw data from each of these databases are described in this section. Following this, we outline the two fusion paradigms examined for the binary classification task of T2DM prediction and provide details of their architectures and training methods.

\subsection{Data acquisition}
\begin{table*}[h]
    \caption{EHR features extracted from the MIMIC-IV database.}
    \label{tab: EHRfeats}
    \centering
    \begin{tabular}{cccc}
        \toprule
        Category&Feature&Unit or encoding $^{\mathrm{a}}$&Modelled as\\
        \toprule
        \multirow{4}{*}{Demographic factors}&Age&Years&Static\\
        &Biological sex&Female: 1, Male: 2&Static\\
        &Height&cm&Time series\\
        &Weight&kg&Time series\\
        \midrule
        \multirow{6}{*}{Vital signs}&Diastolic blood pressure&mmHg&Time series\\
        &Heart rate&bpm&Time series\\
        &Respiratory rate&insp/min&Time series\\
        &Systolic blood pressure&mmHg&Time series\\
        &Temperature&\(^{\circ}\)C&Time series\\
        &Urine output&mL&Time series\\
        \midrule
        Hereditary factors&Family history of T2DM&Yes: 1, No: 0&Static\\
        \bottomrule
        \multicolumn{4}{l}{$^{\mathrm{a}}$ bpm = beats per minute, insp/min = inspirations per minute.}\\
    \end{tabular}
\end{table*}

\subsubsection{EHR data}
MIMIC-IV is a comprehensive longitudinal database containing the EHRs of 299,712 adult patients (aged 18 or older), who were admitted to an Intensive Care Unit (ICU) at the Beth Israel Deaconess Medical Centre (BIDMC) in the USA between 2008 and 2019. All de-identified information relating to patients, billing and diagnosis codes, laboratory tests, and physiological measurements is organised into 31 data tables, all linked by unique patient identifiers. The hierarchical structure of the database progresses from patient information to hospital admissions, and finally to ICU stays.

From this extensive data source, 11 features, including both time series and static variables, are selected based on their established associations with T2DM (refer to Table \ref{tab: EHRfeats}). Demographic factors such as age, biological sex, height, and weight are known risk indicators \cite{M2019Diabetes}. While the likelihood of insulin resistance rises with age and obesity \cite{M2020Implications}, differences in body fat composition (caused by biological sex) act as useful markers of the disease \cite{M2022Mellitus}. A genetic predisposition to T2DM also exists. Additionally, vital signs, such as heart rate, respiratory rate, body temperature, and blood pressure, can reveal cardiovascular abnormalities that often co-occur with T2DM. Lastly, a large SHAP (SHapley Additive exPlanations) value for urine output has also been reported in some studies \cite{M2022Learning}, emphasising its predictive powers.

\subsubsection{CXR images}
The MIMIC-CXR-JPG database includes CXRs and associated metadata from patients admitted to the BIDMC between 2011 and 2016. This allows for the linkage of relevant CXR images to the EHRs available in MIMIC-IV. While the database contains lateral, anteroposterior (AP), and posteroanterior (PA) projections, PA CXRs are selected for this study due to their superior image quality and efficacy in capturing the lungs and bony thorax \cite{M2019Quality}.

\subsubsection{ECG data}
The MIMIC-IV-ECG database is a subset of MIMIC-IV, containing 12-lead, 10-second ECG signals sampled at 500Hz, along with relevant metadata. Since these are diagnostic ECGs, missing values are rare, and baseline wandering has already been removed through high-pass filters, minimising the need for additional signal processing.

\subsection{Data preprocessing}
The datasets are curated such that each input sample corresponds to a unique ICU stay within a patient's specific hospital admission (referred to as an episode), enabling multiple samples to exist for the same patient across different admissions. Each sample is labelled based on the patient's T2DM status. Episodes containing all three modalities after preprocessing form the \textit{$\mathbf{D}_{\boldsymbol{E+C+G}}$} dataset, while the \textit{$\mathbf{D}_{\boldsymbol{E+C}}$} dataset is derived by excluding the ECG modality from these same episodes.

This approach, where subsequent visits are treated as distinct samples, accounts for the substantial changes that could occur in a patient's physiological functions over time. It also promotes resource-efficient clinical practices by allowing CXRs and ECGs to be ordered only when necessary.

\subsubsection{EHR data}
The preprocessing pipeline described in \cite{M2022images}, which was adapted from \cite{I2019data}, is modified to meet the specific requirements of this study. First, admissions without an ICU stay or those involving transfers between ICU and non-ICU wards are excluded. In cases where multiple consecutive ICU stays occur during the same visit, only the initial stay is retained, forming unique visit-stay pairs. Next, vital signs measurements (referred to as events) are filtered to extract the time series features listed in Table \ref{tab: EHRfeats} (refer to Appendix \ref{sec: app_ehr} for specific details). These measurements undergo cleaning, including unit conversions and outlier handling, and any ICU stays without corresponding events are discarded at this stage. In addition to time series data, static features are also retrieved. Among them, the `family history of T2DM' feature is derived from the presence of relevant diagnosis codes (see Table \ref{tab: icd}). Finally, the time series features are sampled at 30-minute intervals for a 48-hour period, with zero imputation implemented to handle missing values. Static features remain constant across all timestamps. This results in \(\mathbf{E}_i\in\mathbb{R}^{96\times11}\), representing the EHR modality for each sample.

Furthermore, during this stage in the preprocessing pipeline, the patients' T2DM status is determined by utilising the ICD (International Classification of Diseases) codes employed in the American healthcare billing system. Specifically, MIMIC-IV incorporates both the ICD-9-CM \cite{M2021ICD-9-CM} and ICD-10-CM \cite{M2024ICD-10-CM} systems. Upon identifying the T2DM-related codes from these systems (refer to Table \ref{tab: icd}), binary labels are assigned to each sample. A positive label is allocated to any sample where one or more of these codes appear in the patient's EHR.

\begin{table}[h!]
    \caption{T2DM-relevant ICD-9-CM and ICD-10-CM codes.}
    \label{tab: icd}
    \centering
    \begin{tabular}{lcc}
        \toprule
        & ICD-9-CM & ICD-10-CM \\
        \midrule
        T2DM $^{\mathrm{a}}$ & 250x0, 250x2 & E11x \\
        Family-history of T2DM & V180 & Z833 \\
        \bottomrule
        \multicolumn{3}{l}{$^{\mathrm{a}}$ x is a placeholder for numeric characters, e.g., ICD-9-CM}\\
        \multicolumn{3}{l}{code 25012 indicates T2DM with ketoacidosis.}\\
    \end{tabular}
\end{table}

\subsubsection{CXR images}
First, to ensure uniformity and reduce computational and memory loads, all CXR images are resized to a width of 384 pixels on the shorter side, while maintaining the original aspect ratio. Next, patients with episodes resulting from the EHR preprocessing step, who had at least one PA CXR performed within a 30-day window around the patient's first hospital admission and final discharge, are identified. Finally, only the initial X-ray within this period is selected and paired with all episodes for the patient, forming their \(\mathbf{C}_i\in\mathbb{R}^{3\times h\times w}\).

\subsubsection{ECG data}
Patients with episodes resulting from the EHR preprocessing step, who had at least one ECG signal recorded between their first hospital admission and final discharge, are identified. Due to the low prevalence of missing values, any ECGs with missing data are discarded. Each patient's first ECG is then paired with all of their episodes.

Following the American Heart Association's recommendations for low and high-frequency filtering \cite{M2007Electrocardiogram}, all ECG leads undergo fifth-order Butterworth filtering with 0.5Hz and 150Hz as the cut-off frequencies. Finally, each ECG signal is transformed into a matrix, \(\mathbf{G}_i\in\mathbb{R}^{100\times12}\) by averaging every 50 rows. This dimension reduction helps align the data structures of the ECG and EHR modalities more closely.

\subsection{Datasets preparation}
The episodes are partitioned randomly into train (70\%), hold-out validation (10\%), and test (20\%) sets. At this stage, standardisation of \(\mathbf{E}_i\) and \(\mathbf{G}_i\) across all partitions is performed using the means ($\mu$) and standard deviations ($\sigma$) computed from the train set samples:
\begin{equation}
    \mathbf{E}_{i(m,n)} = \frac{\mathbf{E}_{i(m,n)} - \mu_n}{\sigma_n}
\end{equation}
Finally, samples in each partition of the \textit{$\mathbf{D}_{\boldsymbol{E+C+G}}$} dataset are represented as $\{(\mathbf{E}_i, \mathbf{C}_i, \mathbf{G}_i, y_i)\,|\,i=1,\ldots,N\}$, while for the \textit{$\mathbf{D}_{\boldsymbol{E+C}}$} dataset, they are $\{(\mathbf{E}_i, \mathbf{C}_i, y_i)\,|\,i=1,\ldots,N\}$, where \(y_i\in\{0,1\}\) indicates the T2DM diagnosis status. Additionally, attention masks, \(E_i\in\mathbb{R}^{11}\) and \(G_i\in\mathbb{R}^{12}\), are generated to facilitate the transformer model's processing. These are binary row vectors, where a zero at a specific index indicates that the corresponding column in \(\mathbf{E}_i\) or \(\mathbf{G}_i\), respectively, is empty.

\subsection{Multimodal fusion paradigms}
\begin{figure*}
    \centering
    \includegraphics[width=0.75\linewidth]{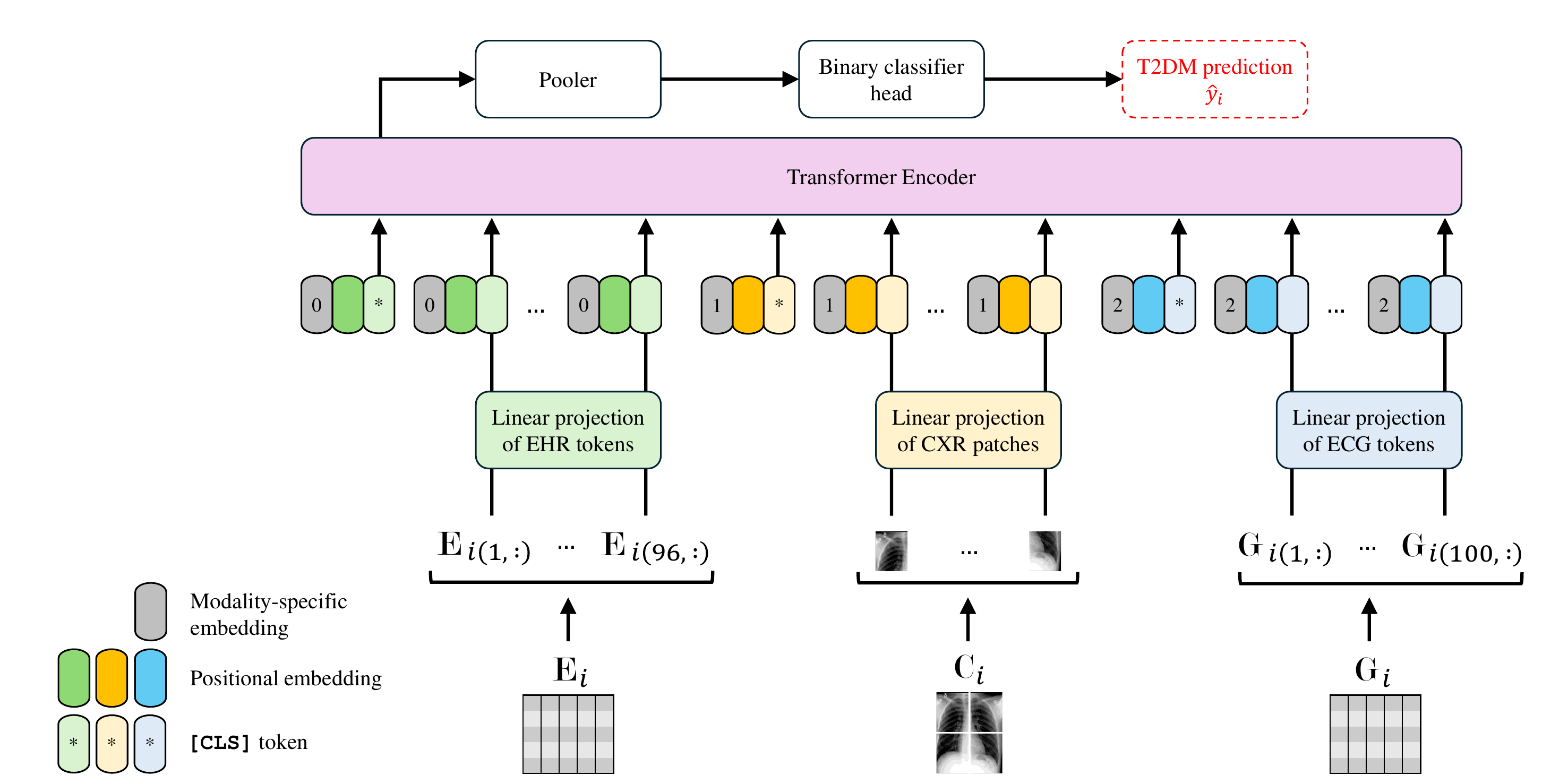}
    \caption{ViLT model architecture with the \textit{$\mathbf{D}_{\boldsymbol{E+C+G}}$} dataset (illustration inspired by \cite{M2021Supervision}). EHR and ECG embeddings are extracted separately and independently from the ViT, while CXR images are processed directly by the ViT, enabling cross-modal interactions.}
    \label{fig: ViLT}
\end{figure*}

\begin{figure*}
    \centering
    \includegraphics[width=0.75\linewidth]{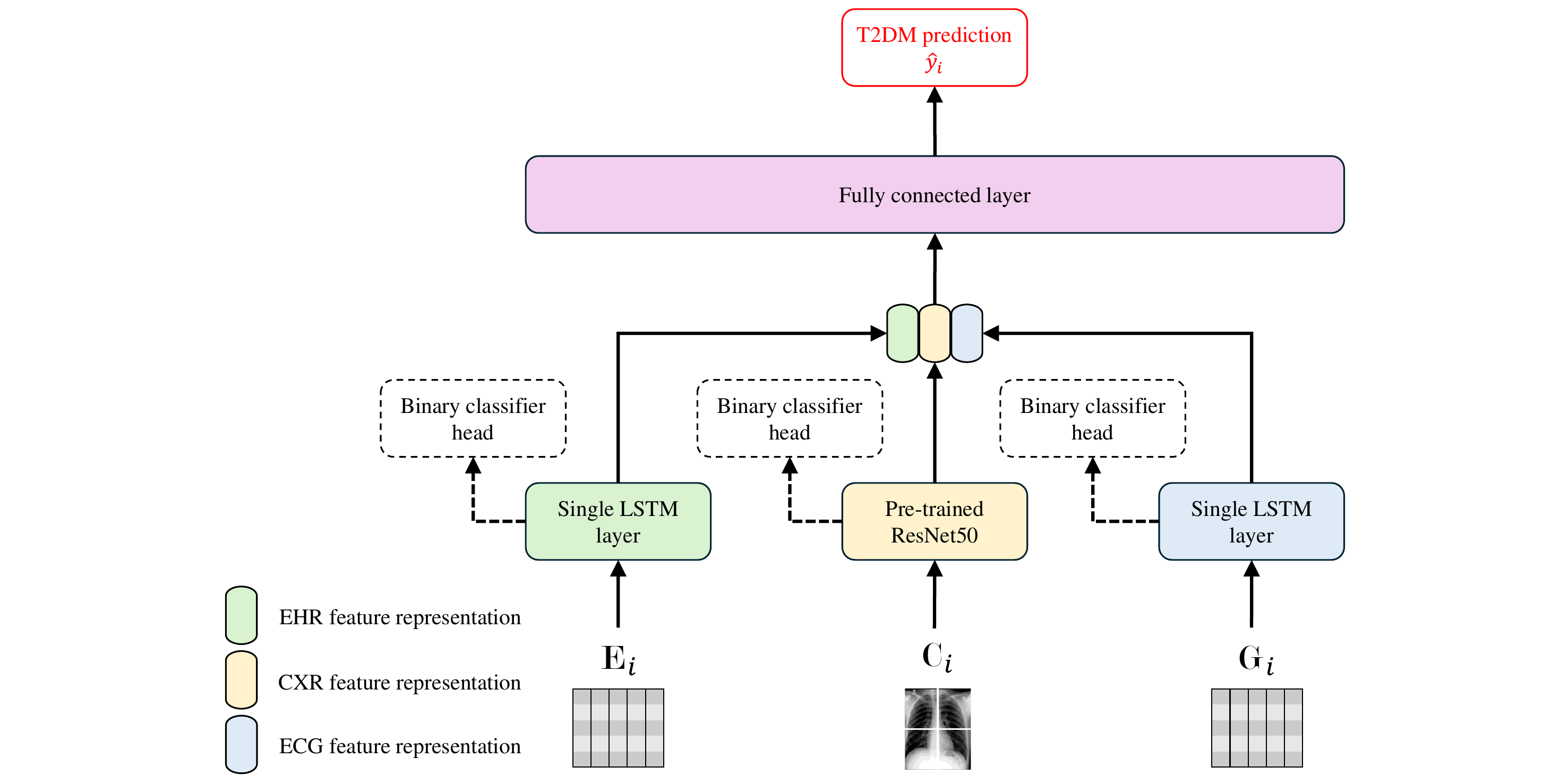}
    \caption{ResNet-LSTM model architecture with the \textit{$\mathbf{D}_{\boldsymbol{E+C+G}}$} dataset (illustration inspired by \cite{M2022images}). The classifier heads attached to the encoders are utilised exclusively in the $\mathbf{ResNet}\text{-}\mathbf{LSTM}_{\boldsymbol{Early}}$ training strategy during its encoder pre-training stage.}
    \label{fig: ResNet-LSTM}
\end{figure*}

\subsubsection{Early fusion: multimodal transformer model}
Despite the enhanced performance achieved by incorporating self-attention mechanisms, many multimodal transformer architectures continue to rely on deep CNN backbones \cite{I2024Review}, such as EfficientNet or ResNet, for embedding images. This results in larger models, increased training expenses, and longer inference times—drawbacks that are particularly disadvantageous in clinical settings where timely decision-making is crucial. Addressing these challenges, this study employs the pre-trained ViLT model \cite{M2021Supervision} (denoted as $\mathbf{ViLT}$), which uses the Vision Transformer (ViT) \cite{M2021Scale} for cross-modal interactions.

The ViT accounts for the non-sequential nature of images by dividing them into fixed-size patches and applying a linear projection, creating input patches that function like tokens in Natural Language Processing. These patches are then passed through a conventional transformer encoder \cite{M2017Need}, generating contextualised representations while simultaneously allowing the ViT to serve as the image encoder. By eliminating the need for a separate convolutional encoder, this approach produces a lightweight model.

Furthermore, ViLT adopts a single-stream, or early fusion, strategy, where all embeddings are concatenated before being input into the transformer encoder. This resembles the decision-making processes of clinical professionals and enables the model to learn from the relationships between different modalities.

The pre-trained ViLT model is adapted for T2DM prediction using transfer learning, where the model, initialised with pre-trained weights, is fine-tuned across all layers. This training strategy is ideal given the scarcity of large, well-labelled, domain-specific multimodal datasets in the medical field, which are typically required to train models from scratch. Additionally, considering ViLT was originally trained on non-clinical data, the use of a generalist biomedical pre-trained transformer \cite{I2024Tasks} might seem preferable. However, research \cite{M2023Phenotyping} has shown that fine-tuning models pre-trained on out-of-domain clinical data yields outcomes comparable to those pre-trained exclusively on non-clinical datasets.

Before fine-tuning, three key modifications are made to the ViLT architecture (see Fig. \ref{fig: ViLT} for an overview):
\begin{enumerate}
    \item {Although both EHR and ECG modalities are modelled as time series, $\mathbf{G}_i$ is introduced as an additional, separate input modality rather than being directly concatenated with $\mathbf{E}_i$. This separation is necessary due to significant differences in their sampling frequencies, which would otherwise require resampling and result in information loss if concatenation were implemented.}
    \item {The EHR and ECG embeddings for each sample $i$, denoted $\overline{\mathbf{E}}_i$ and $\overline{\mathbf{G}}_i$, are generated as detailed below:
        \begin{equation}
            \overline{\mathbf{E}}_i = concat(\mathbf{E}_{CLS},\phantom{.}\mathbf{E}_i\times\mathbf{E}_{proj}) + \mathbf{E}_{pos} + \mathbf{E}_{type}
        \end{equation}
        \begin{equation}
            \overline{\mathbf{G}}_i = concat(\mathbf{G}_{CLS},\phantom{.}\mathbf{G}_i\times\mathbf{G}_{proj}) + \mathbf{G}_{pos} + \mathbf{G}_{type}
        \end{equation}
        
        Specifically, the EHR tokens in $\mathbf{E}_i$ are first linearly projected to match ViLT's embedding dimension ($d_h = 768$) using $\mathbf{E}_{proj}\in\mathbb{R}^{11\times{d_h}}$. Following this, an additional \texttt{[CLS]} token \(\mathbf{E}_{CLS}\in\mathbb{R}^{d_h}\), which is later used for the classification task, is prepended to the tensor. Positional embeddings \(\mathbf{E}_{pos}\in\mathbb{R}^{97\times{d_h}}\), computed as described in \cite{M2017Need}, and modality-specific embeddings \(\mathbf{E}_{type}\in\mathbb{R}^{d_h}\) are added element-wise. Finally, an extra element is also prepended to the mask $E_i$ to account for the \texttt{[CLS]} token, creating $\overline{E}_i$. While an identical embedding process is adopted for the ECG tokens, the CXR embedding $\overline{\mathbf{C}}_i$ and its corresponding mask $\overline{C}_i$ are generated through the linear projection of $32\times32$ flattened image patches, as fully described in \cite{M2021Supervision}, and this process remains unmodified.}
    \item {A binary classifier head is added to predict the labels $\hat{y}_i\in[0,1]$ as part of the downstream classification task, with any additional layers or parameters initialised appropriately.}
\end{enumerate}

\subsubsection{Joint fusion: ResNet-LSTM model}
The joint fusion-based ResNet-LSTM architecture is an adaptation of models typically employed in recent multimodal T2DM studies \cite{I2020Records}\cite{I2022photographs}\cite{I2021images}, which integrate biomedical imaging data with static risk factors from EHRs. While existing approaches utilise CNN-based image encoders, representation learning for the EHR modality is limited to simple linear projections. In this work, however, the sequential nature of EHR and ECG data necessitates the use of recurrent neural networks (RNNs) for encoding these modalities. Specifically, $\mathbf{E}_i$ and $\mathbf{G}_i$ (again treated as distinct inputs) are processed through separate single-layer LSTMs with hidden sizes of 256, while a pre-trained ResNet50 serves as the CXR encoder (see Fig. \ref{fig: ResNet-LSTM} for an overview). Standard image transformations required for fine-tuning ResNet50 are applied to $\mathbf{C}_i$.

The modality-specific encoders are trained using two methods \cite{M2022images} to explore different representation learning settings, yielding two variants of the ResNet-LSTM model:

\begin{enumerate}
    \item Early training (resulting in $\mathbf{ResNet}\text{-}\mathbf{LSTM}_{\boldsymbol{Early}}$): Each encoder is pre-trained individually on its respective modality from the multimodal dataset, with binary classifier heads attached to facilitate this learning. After pre-training, the classifier heads are ignored and the encoder layers are frozen. The final fully connected layer, which classifies using the concatenated modality embeddings, is then trained and validated on the multimodal dataset, completing the modular training process.
    \item Joint training (resulting in $\mathbf{ResNet}\text{-}\mathbf{LSTM}_{\boldsymbol{Joint}}$): The three encoders and the classifier are trained and validated simultaneously with the multimodal dataset, resulting in an end-to-end trained model.
\end{enumerate}

\section{Experiments and Results}
\label{sec: exp+results}
\subsection{Experimental settings}
All experiments are conducted on a single NVIDIA 80Gb A100 GPU using the PyTorch (version 1.13.1) framework and CUDA Toolkit (version 11.7), alongside the PyTorch Lightning (version 1.1.4) and cuDNN (version 8.5.0) wrappers. Test set performance is evaluated on both the \textit{$\mathbf{D}_{\boldsymbol{E+C+G}}$} and \textit{$\mathbf{D}_{\boldsymbol{E+C}}$} datasets using the AUROC (Area Under the Receiver Operating Characteristic Curve), AUPRC (Area Under the Precision-Recall Curve, also referred to as average precision), and accuracy metrics. The 95\% confidence intervals (CI) for these metrics are computed using bootstrapping with 1000 iterations. The hyperparameters explored, based on model performances on the hold-out validation set, included learning rate (\{$10^{-3}$, $10^{-4}$, $10^{-5}$\}) and batch size (\{64, 128, 256, 512\}), as well as sampling rate (every \{15, 30, 60\} minutes), sampling duration (\{24, 48, 72\} hours), and imputation strategy (\{zero, mean, previous, next\}) for the EHR features. Following hyperparameter tuning, a sampling rate of every 30 minutes for 48 hours and zero imputation are chosen for the EHR modality, with additional model-specific details summarised below.

\subsubsection{Early fusion: multimodal transformer model}
The ViLT model is fine-tuned using the AdamW optimiser, which minimises the binary cross-entropy loss, computed as
\begin{equation}
    \label{eq: bce}
    L(y,\hat{y}) = -\frac{1}{B}\sum^{B}_{i=1} (y_i \log\hat{y_i} + (1-y_i) \log(1-\hat{y_i})),
\end{equation}
where $B$ is the batch size. The model is trained for a maximum of 20 epochs, with two validation epochs per training epoch. Early stopping is employed to prevent overfitting, triggered when the AUROC metric on the hold-out validation set shows no improvement over 2 consecutive training epochs. A learning rate of $10^{-4}$, with a linear warmup and decay schedule, and a batch size of 256 are selected after hyperparameter tuning.

\subsubsection{Joint fusion: ResNet-LSTM model}
All stages across both training methods utilised the Adam optimiser to minimise the loss in (\ref{eq: bce}), with an optimal batch size of 64 and a maximum of 50 training epochs. While the same validation epoch pattern is implemented, early stopping is set to a patience level of 5 training epochs. The learning rate remains at $10^{-4}$, but with no warmup and a reduction by a factor of 0.5 whenever the validation loss ceases to reduce for 5 consecutive training epochs.

\subsection{Main results}
\subsubsection{Datasets}
Preprocessing of the MIMIC-IV database yielded 61,911 episodes across 48,232 patients. Among these, 16,300 episodes from 9953 patients had an associated CXR image, and 45,404 episodes from 32,464 patients had a corresponding ECG signal. Ultimately, the \textit{$\mathbf{D}_{\boldsymbol{E+C+G}}$} and \textit{$\mathbf{D}_{\boldsymbol{E+C}}$} datasets both contain 14,091 episodes linked to 7861 unique patients.

\begin{table}[t]
    \caption{Baseline characteristics of samples in the datasets, computed using the static EHR features.}
    \label{tab: dataset}
    \centering
    \begin{tabular}{lccc}
    \toprule
    Feature $^{\mathrm{a}}$ & \makecell[c]{Train set \\ ($N = 9863$)} & \makecell[c]{Validation set $^{\mathrm{b}}$ \\ ($N = 1409$)} & \makecell[c]{Test set \\ ($N = 2819$)}\\
    \midrule
    Age & 61.46 $\pm$ 15.62 & 61.46 $\pm$ 15.61 & 61.31 $\pm$ 15.69 \\
    \phantom{xx}18-29 & 364 (3.69\%) & 45 (3.19\%) & 101 (3.58\%) \\
    \phantom{xx}30-39 & 565 (5.73\%) & 88 (6.25\%) & 160 (5.68\%) \\
    \phantom{xx}40-49 & 1150 (11.66\%) & 176 (12.49\%) & 347 (12.31\%) \\
    \phantom{xx}50-59 & 2076 (21.05\%) & 281 (19.94\%) & 606 (21.50\%) \\
    \phantom{xx}60-69 & 2504 (25.39\%) & 362 (25.69\%) & 717 (25.43\%) \\
    \phantom{xx}70-79 & 1918 (19.44\%) & 273 (19.38\%) & 507 (17.98\%) \\
    \phantom{xx}80-89 & 1128 (11.44\%) & 163 (11.57\%) & 334 (11.85\%) \\
    \phantom{xx}90-99 & 158 (1.60\%) & 21 (1.49\%) & 47 (1.67\%) \\
    \midrule
    \multicolumn{4}{l}{Biological sex} \\
    \phantom{xx}Male & 5527 (56.04\%) & 809 (57.42\%) & 1542 (54.70\%) \\
    \phantom{xx}Female & 4336 (43.96\%) & 600 (42.58\%) & 1277 (45.30\%) \\
    \midrule
    Family history & \multirow{2}{*}{66 (0.70\%)} & \multirow{2}{*}{9 (0.64\%)} & \multirow{2}{*}{18 (0.64\%)} \\
    of T2DM & & & \\
    \midrule
    T2DM & 3160 (32.04\%) & 438 (31.09\%) & 836 (29.66\%) \\
    \bottomrule
    \multicolumn{4}{l}{$^{\mathrm{a}}$ Characteristics are reported as either \{mean $\pm$ standard deviation\} or} \\
    \multicolumn{4}{l}{\{count (\%)\}} \\
    \multicolumn{4}{l}{$^{\mathrm{b}}$ Hold-out validation set} \\
    \end{tabular}
\end{table}

\begin{table*}[h]
    \caption{Test set performance of DL models trained on both multimodal datasets.}
    \label{tab: main_results}
    \centering
    \begin{tabular}{cccc}
    \toprule
    \multirow{2}{*}[-0.7ex]{Model} & \multirow{2}{*}[-0.7ex]{Metric} & \multicolumn{2}{c}{Dataset} \\
    \cmidrule(r){3-4}
    & & \multicolumn{1}{c}{\textit{$\mathbf{D}_{\boldsymbol{E+C+G}}$}} & \multicolumn{1}{c}{\textit{$\mathbf{D}_{\boldsymbol{E+C}}$}} \\
    \midrule
    \multirow{3}{*}{$\mathbf{ViLT}$}  & AUROC    & 0.8481 (0.8320, 0.8648) & 0.8470 (0.8306, 0.8642) \\
                                      & AUPRC    & 0.7325 (0.7010, 0.7645) & 0.7272 (0.6930, 0.7590) \\
                                      & Accuracy & 0.8233 (0.8102, 0.8379) & 0.8209 (0.8070, 0.8343) \\
    \midrule
    \multirow{3}{*}{$\mathbf{ResNet}\text{-}\mathbf{LSTM}_{\boldsymbol{Early}}$} & AUROC & 0.8576 (0.8418, 0.8741) & 0.8578 (0.8420, 0.8742) \\
                                      & AUPRC    & 0.7565 (0.7294, 0.7861) & 0.7588 (0.7323, 0.7884) \\
                                      & Accuracy & 0.8339 (0.8219, 0.8474) & 0.8357 (0.8229, 0.8499) \\
    \midrule
    \multirow{3}{*}{$\mathbf{ResNet}\text{-}\mathbf{LSTM}_{\boldsymbol{Joint}}$} & AUROC & 0.8616 (0.8469, 0.8757) & 0.8592 (0.8425, 0.8751) \\
                                      & AUPRC    & 0.7503 (0.7201, 0.7812) & 0.7562 (0.7260, 0.7859) \\
                                      & Accuracy & 0.8236 (0.8108, 0.8378) & 0.8197 (0.8059, 0.8332) \\
    \bottomrule
    \end{tabular}
\end{table*}

\begin{table*}[h]
    \caption{Pertinent AUROC scores from ablation studies.}
    \label{tab: abl_results}
    \centering
    \begin{tabular}{cccc}
    \toprule
    \multirow{2}{*}[-0.7ex]{Model} & \multirow{2}{*}[-0.7ex]{Ablation} & \multicolumn{2}{c}{Dataset} \\
    \cmidrule(r){3-4}
    & & \multicolumn{1}{c}{\textit{$\mathbf{D}_{\boldsymbol{E+C+G}}$}} & \multicolumn{1}{c}{\textit{$\mathbf{D}_{\boldsymbol{E+C}}$}} \\
    \midrule
    \multirow{4}{*}{$\mathbf{ViLT}$} & Lack of pre-training & 0.7223 (0.7030, 0.7432) & 0.6909 (0.6701, 0.7126) \\
    & Noisy inputs & 0.5779 (0.5564, 0.6007) & 0.6189 (0.5961, 0.6416) \\
    & Missing CXR modality & 0.7248 (0.7058, 0.7449) & 0.7319 (0.7125, 0.7510) \\
    & – & 0.8481 (0.8320, 0.8648) & 0.8470 (0.8306, 0.8642) \\
    \midrule
    \multirow{4}{*}{$\mathbf{ResNet}\text{-}\mathbf{LSTM}_{\boldsymbol{Early}}$} & Lack of pre-training & 0.8373 (0.8211, 0.8546) & 0.8366 (0.8203, 0.8537) \\
    & Noisy inputs & 0.4862 (0.4639, 0.5093) & 0.4924 (0.4695, 0.5157) \\
    & Missing CXR modality & 0.7617 (0.7429, 0.7806) & 0.7441 (0.7256, 0.7631) \\
    & – & 0.8576 (0.8418, 0.8741) & 0.8578 (0.8420, 0.8742) \\
    \midrule
    \multirow{4}{*}{$\mathbf{ResNet}\text{-}\mathbf{LSTM}_{\boldsymbol{Joint}}$} & Lack of pre-training & 0.8566 (0.8427, 0.8723) & 0.8405 (0.8243, 0.8567) \\
    & Noisy inputs & 0.5220 (0.4988, 0.5448) & 0.5773 (0.5543, 0.5990) \\
    & Missing CXR modality & 0.7629 (0.7446, 0.7813) & 0.7426 (0.7233, 0.7624) \\
    & – & 0.8616 (0.8469, 0.8757) & 0.8592 (0.8425, 0.8751) \\
    \bottomrule
    \end{tabular}
\end{table*}

The baseline characteristics of the dataset partitions, based solely on the static EHR features, are summarised in Table \ref{tab: dataset}. They highlight an age distribution predominantly skewed toward older individuals aged 50 and 79, which aligns with the typical onset of T2DM during these years. However, younger populations are also represented, with 9.39\% of the total dataset comprising individuals under the age of 40. The inclusion of this younger cohort prevents age-related bias, allowing the model to capture early signs of the disease. Moreover, together with a well-balanced sex distribution across all partitions, it enables the models to effectively consider both high- and low-risk individuals. The relatively high prevalence of T2DM-positive samples (31.47\% overall) suggests no significant dataset imbalance problems, thereby facilitating stable model convergence.

\subsubsection{Multimodal fusion paradigms}
The metrics achieved on the test sets of the two multimodal datasets by the multimodal transformer and both variants of the ResNet-LSTM model are detailed in Table \ref{tab: main_results}.

While all classifiers achieved commendable results, $\mathbf{ResNet}\text{-}\mathbf{LSTM}_{\boldsymbol{Joint}}$ demonstrated the best performance, with an AUROC of 0.8616 (95\% CI: 0.8469, 0.8757) and 0.8592 (95\% CI: 0.8425, 0.8751) on the \textit{$\mathbf{D}_{\boldsymbol{E+C+G}}$} and \textit{$\mathbf{D}_{\boldsymbol{E+C}}$} datasets, respectively. Its improvement over $\mathbf{ResNet}\text{-}\mathbf{LSTM}_{\boldsymbol{Early}}$ is speculated to derive from the multimodal context provided during end-to-end training. This approach enables better generalisation by dynamically exploiting cross-modal relationships instead of prematurely freezing the modality encoders.

On the other hand, the lower performance of $\mathbf{ViLT}$ can be attributed to two potential limitations. First, despite the implementation of regularisation techniques such as weight decay, large models like ViLT tend to overfit quickly, particularly when fine-tuned on relatively small datasets, which, in this case, are also out-of-domain. Second, although the absence of a deep convolutional backbone offers certain benefits, the patch-based representations used by ViT likely hinder the learning of local and fine-grained features in CXRs, which may be crucial for detecting T2DM. This observation justifies the frequent use of CNN-based representation learning for biomedical images and encourages the exploration of alternative multimodal transformer architectures for this task.

Furthermore, while the inclusion of ECG data provided some complementary diagnostic value, its minimal impact suggests that the vital sign features within the EHR data alone may be sufficient for assessing patients' cardiovascular health.

Finally, a comparison with the baseline CXR-only DL model \cite{I2023radiographs}, which achieved an AUROC of 0.84 (95\% CI: 0.83, 0.85) after training on 271,065 CXRs, confirms the added diagnostic value of combining EHR and CXR data in conjunction with end-to-end DL models. Specifically, the $\mathbf{ResNet}\text{-}\mathbf{LSTM}_{\boldsymbol{Joint}}$ model provides a 2.286\% improvement in AUROC. Moreover, while the baseline model was trained on a significantly larger number of CXRs, this study employed only a fraction of that amount (9863 train set samples and 1409 hold-out validation set samples), making the training process more efficient. This efficiency is particularly advantageous for clinical deployment, as large, well-labelled datasets are scarce in this domain.

\subsection{Ablation studies}
Several ablations are conducted to evaluate various aspects of the fusion paradigms and models used. The relevant AUROC scores are summarised in Table \ref{tab: abl_results}, with the complete set of results provided in Appendix \ref{sec: app_abl}.

\subsubsection{Lack of pre-training}
\label{sec: abl_pretraining}
To understand the prominence of pre-training in transformer-based models and assess their suitability for data-constrained scenarios, the ViT in $\mathbf{ViLT}$ is randomly initialised and trained from scratch. Since its role is primarily associated with the imaging modality, the ResNet50 encoder in the ResNet-LSTM models is trained similarly. The resulting AUROCs on the test sets are presented in Table \ref{tab: abl_results}.

The findings confirm that complex models like ViLT substantially benefit from prior knowledge acquired through pre-training on large, albeit non-clinical, multimodal datasets. Meanwhile, ResNet-LSTM models are negligibly impacted, with end-to-end training offering better compensation for the absence of pre-trained weights through dynamic representation learning.

\subsubsection{Noisy inputs}
\label{sec: abl_noisy}
Robustness against noisy data is critical in medical settings, where motion artefacts and data collection errors are inevitable. To evaluate this, Gaussian, Poisson, and uniform noise with amplitudes of 0.1, 0.5, and 0.7 are added to all test set samples' $\mathbf{E}_i$ and $\mathbf{G}_i$. Additionally, Gaussian, Poisson and Salt-and-Pepper noise with the same amplitudes are introduced to their $\mathbf{C}_i$. The average AUROC scores across these noise levels are reported in Table \ref{tab: abl_results}.

While significant performance declines are observed with the addition of noise, they are more prominent in joint fusion architectures, with decreases of 43.31\% and 39.42\% for the early and joint training strategies, respectively. The superior robustness of $\mathbf{ViLT}$ validates the ability of early fusion to exploit inter-modal interactions more effectively. This robustness can be further attributed to the ViT's enhanced generalisation capabilities compared to CNNs. By leveraging self-attention to capture global-level information across all modalities, the ViT can better handle input corruptions and has been empirically shown to exhibit noise-absorbing capabilities \cite{E2022Transformers}.

\subsubsection{Missing CXR modality}
\label{sec: abl_missing}
To simulate varying levels of missing CXR data, $\mathbf{C}_i$ is replaced with zeros in 30\%, 50\%, and 70\% of the test set samples, and the average AUROC scores across these missing ratios are listed in Table \ref{tab: abl_results}.

The results reveal the resilience of the joint fusion models to sparse data, which is speculated to correlate to their modular architecture. This design enables them to mitigate the impact of a missing modality while still producing robust feature representations from the available data. In contrast, the early fusion architecture is disadvantaged in such scenarios. Additionally, the attention-based modality interaction of $\mathbf{ViLT}$ is likely disrupted by the presence of an all-white image, resulting in poorer performance. This ablation study also clearly demonstrates the diagnostic utility of CXRs.

\section{Conclusion}
\label{sec: conclusion}
To conclude, this paper investigated the efficacy of CXRs, particularly in combination with EHRs, for noninvasive screening of T2DM. We evaluated two end-to-end DL architectures with distinct fusion approaches on multimodal datasets specifically curated for T2DM from the publicly available MIMIC-IV databases. All implemented models outperformed the state-of-the-art CXR-only DL classifier, with the most notable model achieving a 2.3\% increase in performance. These results demonstrate the feasibility of integrating these clinical modalities to enhance early detection of T2DM, even among individuals who may not yet exhibit typical symptoms or risk factors, offering significant benefits for resource-deficient regions. However, the incorporation of ECG data yielded only modest improvements in the predictive capabilities of the models.

Given the critical nature of clinical diagnostics, further validation using external datasets is essential and forms a key direction for future work. This will also help assess the impact of potential limitations in the study's datasets, particularly the approach of treating subsequent ICU stays of a patient as separate samples, despite sharing the same CXR and ECG data. While this methodology improves the efficiency of medical procedures, it may unintentionally bias the model towards chronic conditions captured by these specific modalities.

\newpage
\appendices
\section{EHR features from MIMIC-IV database}
\label{sec: app_ehr}
Table \ref{tab: app_ehr} provides details of the relevant event variables in MIMIC-IV used in this study.
\begin{table}[H]
    \caption{EHR features extracted from the MIMIC-IV database with their corresponding event variables and locations.}
    \label{tab: app_ehr}
    \centering
    \scriptsize
    \begin{tabular}{lll}
    \toprule
    EHR feature&MIMIC-IV event variable& MIMIC-IV table\\
    \midrule
    \multirow{2}{*}{Height}                   & Height                                & chartevents    \\
                                              & Height (cm)                           & chartevents    \\
    \midrule
    \multirow{2}{*}{Weight}                   & Admission Weight   (Kg)               & chartevents    \\
                                              & Admission Weight (lbs.)               & chartevents    \\
                                              \midrule
    \multirow{5}{*}{\makecell[l]{Diastolic blood \\ pressure}}&Arterial Blood Pressure diastolic&chartevents\\
    & ART BP Diastolic& chartevents\\
    & Non Invasive Blood Pressure diastolic & chartevents    \\
    & Manual Blood Pressure Diastolic Left  & chartevents    \\
    & Manual Blood Pressure Diastolic Right & chartevents    \\
    \midrule
    Heart rate                                & Heart rate                            & chartevents    \\
    \midrule
    \multirow{3}{*}{Respiratory rate}         & Respiratory Rate                      & chartevents    \\
                                              & Respiratory Rate (spontaneous)        & chartevents    \\
                                              & Respiratory Rate (Total)              & chartevents    \\
                                              \midrule
    \multirow{5}{*}{\makecell[l]{Systolic blood \\ pressure}}  & Arterial Blood   Pressure systolic    & chartevents    \\
                                              & ART BP Systolic                       & chartevents    \\
                                              & Non Invasive Blood Pressure systolic  & chartevents    \\
                                              & Manual Blood Pressure Systolic Left   & chartevents    \\
                                              & Manual Blood Pressure Systolic Right  & chartevents    \\
                                              \midrule
    \multirow{2}{*}{Temperature}              & Temperature Celsius                   & chartevents    \\
                                              & Temperature Fahrenheit                & chartevents    \\
                                              \midrule
    \multirow{12}{*}{Urine output}            & R Ureteral Stent                      & outputevents   \\
                                              & L Ureteral Stent                      & outputevents   \\
                                              & Foley                                 & outputevents   \\
                                              & Void                                  & outputevents   \\
                                              & Condom Cath                           & outputevents   \\
                                              & Suprapubic                            & outputevents   \\
                                              & R Nephrostomy                         & outputevents   \\
                                              & L Nephrostomy                         & outputevents   \\
                                              & Straight Cath                         & outputevents   \\
                                              & Ileoconduit                           & outputevents   \\
                                              & GU Irrigant Volume In                 & outputevents   \\
                                              & GU Irrigant/Urine Volume Out          & outputevents   \\
    \bottomrule
    \end{tabular}
\end{table}

\section{Complete ablation study results}
\label{sec: app_abl}
The complete set of performance metrics from ablation studies \#1, \#2, and \#3 are reported in Table \ref{tab: app_pretraining}, Tables \ref{tab: app_noisy1} and \ref{tab: app_noisy2}, and Tables \ref{tab: app_missing1} and \ref{tab: app_missing2}, respectively.

\begin{table}[H]
    \caption{All results from Ablation study \#1: Lack of pre-training.}
    \label{tab: app_pretraining}
    \centering
    \scriptsize
    \setlength{\tabcolsep}{3pt}
    \begin{tabular}{cccc}
    \toprule
    \multirow{2}{*}[-0.7ex]{Model} & \multirow{2}{*}[-0.7ex]{Metric} & \multicolumn{2}{c}{Dataset} \\
    \cmidrule(r){3-4}
    & & \multicolumn{1}{c}{\textit{$\mathbf{D}_{\boldsymbol{E+C+G}}$}} & \multicolumn{1}{c}{\textit{$\mathbf{D}_{\boldsymbol{E+C}}$}} \\
    \midrule
    \multirow{3}{*}{$\mathbf{ViLT}$}  & AUROC    & 0.7223 (0.7030, 0.7432) & 0.6909 (0.6701, 0.7126) \\
                                      & AUPRC    & 0.5085 (0.4757, 0.5443) & 0.4624 (0.4308, 0.4962) \\
                                      & Accuracy & 0.7002 (0.6829, 0.7173) & 0.6747 (0.6577, 0.6924) \\
    \midrule
    \multirow{3}{*}{\makecell{\hspace{-2em}$\mathbf{ResNet}$\\\hspace{2em}$\text{-}\mathbf{LSTM}_{\boldsymbol{Early}}$}} & AUROC & 0.8373 (0.8211, 0.8546) & 0.8366 (0.8203, 0.8537) \\
                                      & AUPRC    & 0.7306 (0.7007, 0.7625) & 0.7298 (0.6989, 0.7603) \\
                                      & Accuracy & 0.8070 (0.7928, 0.8208) & 0.8190 (0.8059, 0.8329) \\
    \midrule
    \multirow{3}{*}{\makecell{\hspace{-2em}$\mathbf{ResNet}$\\\hspace{2em}$\text{-}\mathbf{LSTM}_{\boldsymbol{Joint}}$}} & AUROC & 0.8566 (0.8427, 0.8723) & 0.8405 (0.8243, 0.8567) \\
                                      & AUPRC    & 0.7423 (0.7128, 0.7738) & 0.7247 (0.6937, 0.7590) \\
                                      & Accuracy & 0.8077 (0.7942, 0.8215) & 0.7960 (0.7811, 0.8098) \\
    \bottomrule
    \end{tabular}
\end{table}

\begin{table*}[hbtp]
    \caption{Results from Ablation study \#2: Noisy inputs with the $\mathbf{D}_{\boldsymbol{E+C}}$ dataset.}
    \label{tab: app_noisy1}
    \centering
    \scriptsize
    \begin{tabular}{ccccc}
    \toprule
    \multirow{2}{*}[-0.7ex]{Model} & \multirow{2}{*}[-0.7ex]{Metric} & \multicolumn{3}{c}{Noise level} \\
    \cmidrule(r){3-5}
    & & \multicolumn{1}{c}{0.1} & \multicolumn{1}{c}{0.5} & \multicolumn{1}{c}{0.7} \\
    \midrule
    \multirow{3}{*}{$\mathbf{ViLT}$}  & AUROC & 0.6327 (0.6105, 0.6539) & 0.6130 (0.5893, 0.6351) & 0.6112 (0.5885, 0.6357) \\
    & AUPRC & 0.4149 (0.3845, 0.4538) & 0.3893 (0.3617, 0.4237) & 0.3908 (0.3605, 0.4246) \\
    & Accuracy & 0.6999 (0.6832, 0.7180) & 0.7020 (0.6864, 0.7190) & 0.6992 (0.6825, 0.7169) \\
    \midrule
    \multirow{3}{*}{\makecell[l]{$\mathbf{ResNet-LSTM}_{\boldsymbol{Early}}$}} & AUROC & 0.5607 (0.5375, 0.5840) & 0.4568 (0.4341, 0.4805) & 0.4596 (0.4368, 0.4826) \\
    & AUPRC & 0.3333 (0.3091, 0.3600) & 0.2774 (0.2570, 0.3033) & 0.2792 (0.2573, 0.3016) \\
    & Accuracy & 0.7016 (0.6845, 0.7186) & 0.7033 (0.6859, 0.7207) & 0.7033 (0.6859, 0.7207) \\
    \midrule
    \multirow{3}{*}{\makecell[l]{$\mathbf{ResNet-LSTM}_{\boldsymbol{Joint}}$}} & AUROC & 0.5730 (0.5501, 0.5950) & 0.5902 (0.5681, 0.6104) & 0.5688 (0.5446, 0.5916) \\
    & AUPRC & 0.3372 (0.3129, 0.3658) & 0.3565 (0.3317, 0.3872) & 0.3400 (0.3147, 0.3696) \\
    & Accuracy & 0.2970 (0.2796, 0.3144) & 0.2967 (0.2793, 0.3141) & 0.2967 (0.2793, 0.3141) \\
    \bottomrule
    \end{tabular}
\end{table*}

\begin{table*}[hbtp]
    \caption{Results from Ablation study \#2: Noisy inputs with the $\mathbf{D}_{\boldsymbol{E+C+G}}$ dataset.}
    \label{tab: app_noisy2}
    \centering
    \scriptsize
    \begin{tabular}{ccccc}
    \toprule
    \multirow{2}{*}[-0.7ex]{Model} & \multirow{2}{*}[-0.7ex]{Metric} & \multicolumn{3}{c}{Noise level} \\
    \cmidrule(r){3-5}
    & & \multicolumn{1}{c}{0.1} & \multicolumn{1}{c}{0.5} & \multicolumn{1}{c}{0.7} \\
    \midrule
    \multirow{3}{*}{$\mathbf{ViLT}$}  & AUROC & 0.6640 (0.6437, 0.6865) & 0.5325 (0.5105, 0.5552) & 0.5371 (0.5151, 0.5604) \\
    & AUPRC & 0.4312 (0.3996, 0.4640) & 0.3268 (0.3039, 0.3570) & 0.3236 (0.2997, 0.3522) \\
    & Accuracy & 0.7020 (0.6857, 0.7180) & 0.5924 (0.5754, 0.6126) & 0.4867 (0.4683, 0.5055) \\
    \midrule
    \multirow{3}{*}{\makecell[l]{$\mathbf{ResNet-LSTM}_{\boldsymbol{Early}}$}} & AUROC & 0.5360 (0.5133, 0.5587) & 0.4441 (0.4221, 0.4674) & 0.4786 (0.4562, 0.5019) \\
    & AUPRC & 0.3091 (0.2868, 0.3344) & 0.2667 (0.2470, 0.2900) & 0.2916 (0.2700, 0.3187) \\
    & Accuracy & 0.7030 (0.6856, 0.7197) & 0.7033 (0.6859, 0.7207) & 0.7033 (0.6859, 0.7207) \\
    \midrule
    \multirow{3}{*}{\makecell[l]{$\mathbf{ResNet-LSTM}_{\boldsymbol{Joint}}$}} & AUROC & 0.5411 (0.5175, 0.5645) & 0.5167 (0.4950, 0.5396) & 0.5081 (0.4840, 0.5302) \\
    & AUPRC & 0.3151 (0.2932, 0.3416) & 0.3013 (0.2798, 0.3262) & 0.3026 (0.2793, 0.3287) \\
    & Accuracy & 0.7023 (0.6852, 0.7197) & 0.7033 (0.6859, 0.7207) & 0.7033 (0.6859, 0.7207) \\
    \bottomrule
    \end{tabular}
\end{table*}

\begin{table*}[hbtp]
    \caption{Results from Ablation study \#3: Missing CXR modality with the $\mathbf{D}_{\boldsymbol{E+C}}$ dataset.}
    \label{tab: app_missing1}
    \centering
    \scriptsize
    \begin{tabular}{cccccccc}
    \toprule
    \multirow{2}{*}[-0.7ex]{Model} & \multirow{2}{*}[-0.7ex]{Metric} & \multicolumn{3}{c}{Missing ratio} \\
    \cmidrule(r){3-5}
    & & \multicolumn{1}{c}{30\%} & \multicolumn{1}{c}{50\%} & \multicolumn{1}{c}{70\%} \\
    \midrule
    \multirow{3}{*}{$\mathbf{ViLT}$}  & AUROC & 0.7963 (0.7781, 0.8137) & 0.7318 (0.7128, 0.7506) & 0.6675 (0.6465, 0.6887) \\
    & AUPRC & 0.6422 (0.6083, 0.6792) & 0.5586 (0.5253, 0.5958) & 0.4823 (0.4503, 0.5171) \\
    & Accuracy & 0.7801 (0.7659, 0.7950) & 0.7513 (0.7350, 0.7673) & 0.7322 (0.7151, 0.7488) \\ 
    \midrule
    \multirow{3}{*}{\makecell[l]{$\mathbf{ResNet-LSTM}_{\boldsymbol{Early}}$}} & AUROC & 0.8045 (0.7879, 0.8224) & 0.7456 (0.7274, 0.7648) & 0.6821 (0.6614, 0.7022) \\
    & AUPRC & 0.6841 (0.6545, 0.7160) & 0.6005 (0.5672, 0.6340) & 0.5164 (0.4843, 0.5527) \\
    & Accuracy & 0.7960 (0.7807, 0.8112) & 0.7651 (0.7484, 0.7807) & 0.7392 (0.7221, 0.7555) \\ 
    \midrule
    \multirow{3}{*}{\makecell[l]{$\mathbf{ResNet-LSTM}_{\boldsymbol{Joint}}$}} & AUROC & 0.8023 (0.7850, 0.8199) & 0.7444 (0.7254, 0.7658) & 0.6812 (0.6596, 0.7014) \\
    & AUPRC & 0.6734 (0.6415, 0.7084) & 0.5945 (0.5600, 0.6283) & 0.5085 (0.4758, 0.5417) \\
    & Accuracy & 0.6891 (0.6725, 0.7076) & 0.5937 (0.5774, 0.6107) & 0.5018 (0.4851, 0.5209) \\ 
    \bottomrule
    \end{tabular}
\end{table*}

\begin{table*}[hbtp]
    \caption{Results from Ablation study \#3: Missing CXR modality with the $\mathbf{D}_{\boldsymbol{E+C+G}}$ dataset.}
    \label{tab: app_missing2}
    \centering
    \scriptsize
    \begin{tabular}{cccccccc}
    \toprule
    \multirow{2}{*}[-0.7ex]{Model} & \multirow{2}{*}[-0.7ex]{Metric} & \multicolumn{3}{c}{Missing ratio} \\
    \cmidrule(r){3-5}
    & & \multicolumn{1}{c}{30\%} & \multicolumn{1}{c}{50\%} & \multicolumn{1}{c}{70\%} \\
    \midrule
    \multirow{3}{*}{$\mathbf{ViLT}$}  & AUROC & 0.7935 (0.7761, 0.8117) & 0.7272 (0.7084, 0.7471) & 0.6537 (0.6328, 0.6760) \\
    & AUPRC & 0.6452 (0.6140, 0.6811) & 0.5612 (0.5274, 0.5981) & 0.4806 (0.4476, 0.5154) \\
    & Accuracy & 0.7772 (0.7623, 0.7925) & 0.7453 (0.7286, 0.7620) & 0.7258 (0.7091, 0.7418) \\ 
    \midrule
    \multirow{3}{*}{\makecell[l]{$\mathbf{ResNet-LSTM}_{\boldsymbol{Early}}$}} & AUROC & 0.8094 (0.7907, 0.8273) & 0.7647 (0.7463, 0.7838) & 0.7111 (0.6917, 0.7308) \\
    & AUPRC & 0.6759 (0.6436, 0.7091) & 0.6193 (0.5870, 0.6528) & 0.5371 (0.5053, 0.5729) \\
    & Accuracy & 0.7913 (0.7775, 0.8066) & 0.7686 (0.7530, 0.7842) & 0.7452 (0.7296, 0.7615) \\ 
    \midrule
    \multirow{3}{*}{\makecell[l]{$\mathbf{ResNet-LSTM}_{\boldsymbol{Joint}}$}} & AUROC & 0.8141 (0.7977, 0.8317) & 0.7624 (0.7446, 0.7805) & 0.7123 (0.6917, 0.7317) \\
    & AUPRC & 0.6686 (0.6369, 0.7037) & 0.6030 (0.5678, 0.6369) & 0.5232 (0.4897, 0.5599) \\
    & Accuracy & 0.7385 (0.7225, 0.7552) & 0.6678 (0.6501, 0.6842) & 0.6128 (0.5965, 0.6309) \\ 
    \bottomrule
    \end{tabular}
\end{table*}

\clearpage
\bibliographystyle{IEEEtran}
\bibliography{ms}

\EOD
\end{document}